\documentclass{article}

\usepackage{microtype}
\usepackage{subcaption,graphicx}

\usepackage{booktabs} % for professional tables

\usepackage{amsmath,amsfonts,bm}

\def\eqref#1{equation~\ref{#1}}
\def\1{\bm{1}}

\DeclareMathAlphabet{\mathsfit}{\encodingdefault}{\sfdefault}{m}{sl}
\SetMathAlphabet{\mathsfit}{bold}{\encodingdefault}{\sfdefault}{bx}{n}

\newcommand{\pdata}{p_{\rm{data}}}
\newcommand{\xpos}{x_+}
\newcommand{\xneg}{x_-}

\newcommand{\modelname}{{\textsc{CIM}}}

\usepackage[colorlinks=true,linkcolor=blue,citecolor=black]{hyperref}
\usepackage{url}
\usepackage{graphicx}
\usepackage{multicol}
\usepackage{multirow}
\usepackage{booktabs}

\usepackage{wrapfig}
\usepackage{float}
\newcommand{\STAB}[1]{\begin{tabular}{@{}c@{}}#1\end{tabular}}
\usepackage{pifont}
\usepackage{xcolor}
\usepackage{tabularx}
\usepackage[export]{adjustbox}
\usepackage{caption}
\usepackage{wrapfig}
\usepackage{fixltx2e}

\newcommand{\cmark}{\textcolor{green}{\ding{51}}}
\newcommand{\xmark}{\textcolor{red!80!black}{\ding{55}}}

\usepackage{hyperref}

\usepackage[accepted]{icml2021}

\icmltitlerunning{Robust Representation Learning via Perceptual Similarity Metrics}

\begin{document}

\twocolumn[
\icmltitle{Robust Representation Learning via Perceptual Similarity Metrics}

% \icmltitle{Learning Task-Relevant Features via Contrastive Input Morphing}



\icmlsetsymbol{equal}{*}

\begin{icmlauthorlist}
\icmlauthor{Saeid Asgari Taghanaki}{equal,adsk}
\icmlauthor{Kristy Choi}{equal,stfd}
\icmlauthor{Amir Khasahmadi}{adsk}
\icmlauthor{Anirudh Goyal}{mila}
% \icmlauthor{Fiuea Rrrr}{to}
\end{icmlauthorlist}

\icmlaffiliation{adsk}{Autodesk AI Lab}
\icmlaffiliation{stfd}{Computer Science, Stanford University}
\icmlaffiliation{mila}{Mila, Université de Montréal}

\icmlcorrespondingauthor{Saeid Asgari Taghanaki}{saeid.asgari.taghanaki@autodesk.com}
% \icmlcorrespondingauthor{Eee Pppp}{ep@eden.co.uk}

\icmlkeywords{Machine Learning, ICML}

\vskip 0.3in
]



\printAffiliationsAndNotice{\icmlEqualContribution} % otherwise use the standard text.

\begin{abstract}
A fundamental challenge in artificial intelligence is learning useful representations of data that yield good performance on a downstream task, without overfitting to spurious input features. Extracting such task-relevant predictive information is particularly difficult for real-world datasets. In this work, we propose Contrastive Input Morphing (CIM), a representation learning framework that learns \textit{input-space transformations} of the data to mitigate the effect of irrelevant input features on downstream performance. 
Our method leverages a perceptual similarity metric via a triplet loss to ensure that the transformation preserves task-relevant information.
Empirically, we demonstrate the efficacy of our approach on tasks which typically suffer from the presence of spurious correlations: classification with nuisance information, out-of-distribution generalization, and preservation of subgroup accuracies. We additionally show that CIM is complementary to other mutual information-based representation learning techniques, and demonstrate that it improves the performance of variational information bottleneck (VIB) when used together. 
\end{abstract}

\section{Introduction}
\label{intro}
\begin{figure*}
     \centering
     \includegraphics[width=0.7\textwidth]{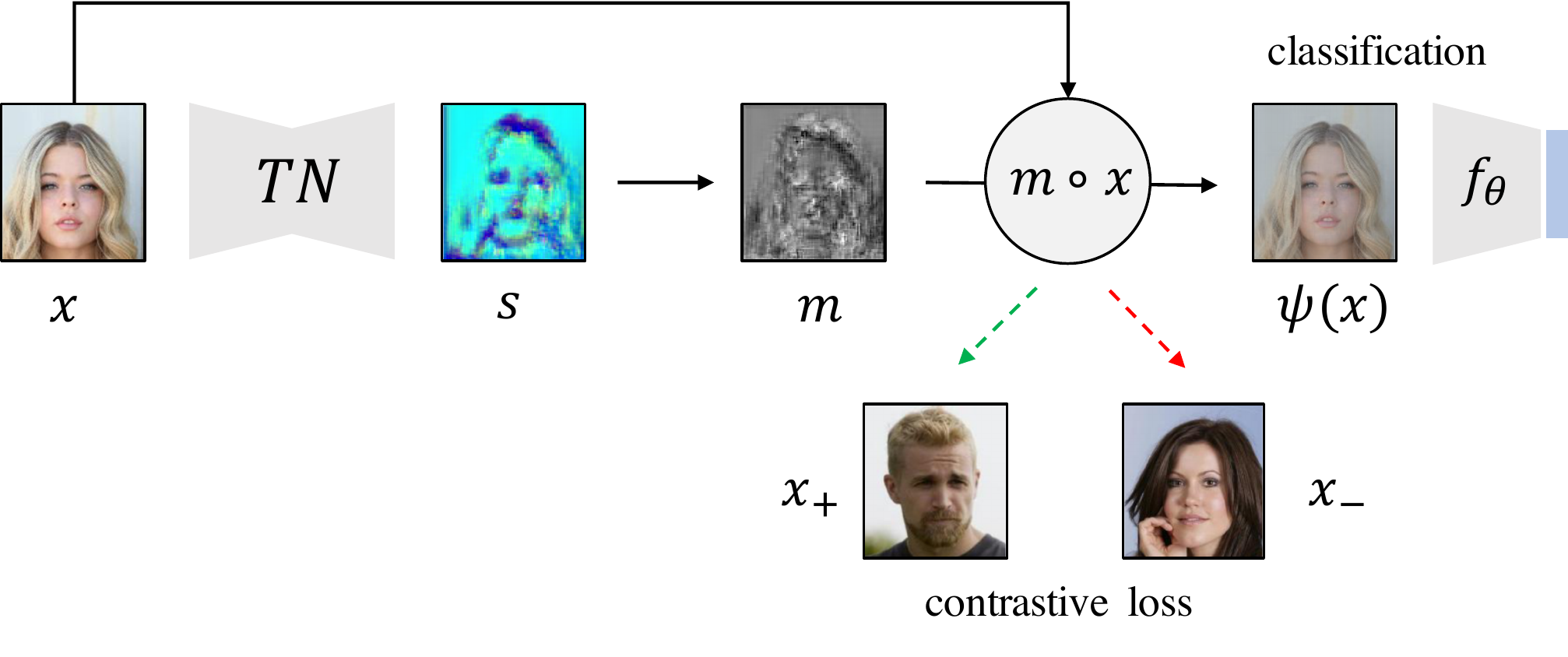}
     \caption{An end-to-end flowchart for the \modelname~training procedure. The Transformation Network (TN) maps an input image $x$ into a weighting matrix $m$, then uses a triplet loss over the modified input using $m$, positive samples, and negative samples to facilitate learning task-relevant representations. The final transformed image $\psi(x) = m \circ x$ is used to train the downstream classifier $f_\theta$.}
     \label{fig:method}
\end{figure*}

At the heart of human intelligence is the property of \textit{robust generalization} -- given one or a handful of examples, we are typically able to learn a concept and apply it across a variety of tasks and conditions.

Modern machine learning aims to replicate this phenomenon

with our artificial agents -- one such way to do so is via representation learning, or extracting features from raw data that enable predictions with high accuracy \citep{hinton2006reducing,vincent2010stacked,chen2016infogan,van2017neural,oord2018representation}. In particular, the recent successes of deep neural networks \citep{dean2012large,lecun2015deep} have been pivotal towards stepping closer to this goal. However, their rapidly growing size and large-scale training procedures, coupled with complex, high-dimensional data sources, pose significant challenges for learning models that perform well on a given task without overfitting to spurious input features~\citep{zhang2016understanding,ilyas2019adversarial,geirhos2020shortcut}. As a result, trained networks have been shown to fail spectacularly  on out-of-domain generalization tasks \citep{beery2018recognition,rosenfeld2018elephant} and exhibit poor performance for rare subgroups present in data \citep{hashimoto2018fairness,goel2020model}, among others. 

A wide range of methods tackle this problem, including regularization, data augmentation, leveraging causal explanations, and self-training \citep{srivastava2014dropout,chen2020self,sagawa2019distributionally,chen2020self,gulrajani2020search}. In particular, prior art places a heavy emphasis on \emph{lossless access} to the input data during training, then constructing a high-level representation which extracts the necessary task relevant information. Yet it is reasonable to assume that in some cases, we desire access to only a \textit{subset} of the input which is relevant to the task -- for example, the background color in an image of a ``7" is unnecessary for identifying its digit class. The fundamental challenge, then, is discerning which parts of the input are relevant without requiring access to \emph{privileged information} (e.g. the nature of the downstream task) at training time.

Our approach, Contrastive Input Morphing (\modelname), leverages labeled supervision to \textit{learn lossy input-space transformations} of the data that mitigate the effect of irrelevant input features on downstream predictive performance. 
The key workhorse of \modelname~is an auxiliary network called the Transformation Network (TN).
Drawing inspiration from the ``robustness" of the human visual system \cite{geirhos2020shortcut}, perceptual similarity metrics \cite{wang2004image,zhang2018unreasonable}, and metric learning \citep{goldberger2004neighbourhood,weinberger2009distance,schroff2015facenet,koch2015siamese}, the TN is trained via a triplet loss that computes the perceptual similarity between sets of transformed inputs, examples from the same class as the input (positive examples), and those from competing classes (negative examples). This measure of similarity is captured by the structural similarity metric, or SSIM \cite{wang2004image} -- a metric developed to extract features of an image that are striking to human perception. Intuitively, this objective uses the shared information from competing classes as a proxy for spurious correlations in the data, while leveraging the shared information within the same class as a heuristic for task-relevancy \citep{khosla2020supervised}.

The framework for \modelname~is quite general: it can be used as a plug-in module for training any downstream classifier, and we demonstrate a particular instance of its compatibility with variational information bottleneck (VIB) \citep{alemi2016deep}, a mutual information (MI)-based representation learning technique. We emphasize that our method does not assume access to the exact nature of the downstream task, such as attribute labels for rare subgroups.

A flowchart of the CIM training procedure can be found in Figure~\ref{fig:method}.

Empirically, we evaluate our method on five different datasets under three settings that suffer from spurious correlations: classification with nuisance background information, out-of-domain (OOD) generalization, and improving accuracy uniformly across subgroups. In the first task, we show that when CIM is used with VIB (CIM+VIB), it outperforms ERM on colored MNIST and improves over the ResNet-50 baseline on the Background Challenge \citep{xiao2020noise}. Similarly, CIM+VIB outperforms relevant baselines using ResNet-18 on the VLCS dataset \citep{torralba2011unbiased} for OOD generalization. For subgroup accuracies, our method outperforms unsupervised methods on CelebA~\citep{liu2015faceattributes} and the Waterbirds dataset \citep{wah2011caltech} in terms of worst-group accuracy.

To the best of our knowledge, this work is the first to explore SSIM in a contrastive learning setup and to demonstrate its usefulness for learning robust representations on tasks which suffer from spurious correlations, unlike previous works which were limited to image classification or adversarial robustness \cite{snell2017learning,abobakr2019ssimlayer}. In summary, our contributions in this work are as follows:
\begin{enumerate}
    \item We propose \modelname, a method demonstrating that lossy access to input data helps extract good task-relevant representations.
    \item We show that \modelname~is complementary to existing methods, as the learned transformations can be leveraged by other MI-based representation learning techniques such as VIB.
    \item We empirically verify the robustness of the learned representations to spurious correlations in the input features on a variety of tasks (Section~\ref{sec:experiments}).
\end{enumerate}
\section{Preliminaries}
\label{setup}

\subsection{Notation and Problem Setup}
We consider the supervised learning setting with inputs $x \in \mathcal{X} \subseteq \mathbb{R}^d$ and corresponding labels $y \in \mathcal{Y} = \{1, \ldots, k\}$. We assume access to samples $\mathcal{D} = \{(x_i, y_i)\}_{i=1}^n$ drawn from an underlying (unknown) distribution $\pdata(x,y)$, and use capital letters to denote random variables, e.g. $X$ and $Y$. We use $P(X,Y)$ to denote their joint distribution as well as $P(\cdot)$ for the marginal distribution.

Our goal is to learn a classifier $f_\theta: \mathcal{X} \rightarrow \mathcal{Y}$, where $f_\theta \in \Theta$ achieves low prediction error according to some loss function $\ell: \Theta \times (\mathcal{X} \times \mathcal{Y}) \rightarrow \mathbb{R}$. Specifically, we minimize:\\
\begin{equation}
    \label{erm}
    \mathcal{L}_{\textrm{sup}}(\theta) = \mathbb{E}_{x,y \sim \pdata(x,y)}[\ell(f_\theta(x), y)] \approx \sum_{i=1}^n \ell(f_\theta(x_i), y_i)
\end{equation}

In addition to good classification performance, we aim to learn representations of the data, which: (a) are highly predictive of the downstream task; and (b) do not rely on spurious input features. That is, the learned representations should be \textit{task-relevant}. 

\subsection{Metric Learning}
Metric learning \cite{goldberger2004neighbourhood} refers to a family of methods which learn a notion of similiarity (the \emph{metric} of interest) between sets of inputs to extract meaningful representations from data. Although several variations of the method exist, they all share a unifying principle in which examples from the same group are closer together in feature space, while those from opposing groups are further apart \cite{davis2007information,weinberger2009distance,schroff2015facenet,koch2015siamese,hoffer2015deep}. Such methods typically operate over a triplet or contrastive loss \cite{khosla2020supervised} which allows for gradient-based learning of an appropriate distance-based measure between the examples.

\subsection{Structural Similarity Metrics}
The key ingredient for training the TN is the SSIM metric \cite{wang2004image}, though we leverage its multi-scale variant (MS-SSIM) \cite{wang2003multiscale} for our experiments. Given a pair of images, SSIM metrics compute local, pixel-wise statistics across their luminance ($l$), contrast ($c$), and structure ($s$) to assign a score for their perceptual image quality. Such comparison functions capture features that the human visual system are sensitive to \cite{hore2010image}, making SSIM a more desirable candidate for comparing images in pixel space relative to others such as the $\ell_2$ distance.

More concretely, for a pair of images $x$ and $y$, we denote the mean pixel intensity of $x$ as $\mu_x$, the standard deviation of its pixel intensity as $\sigma_x$, and the covariance between the pixels in $x$ and $y$ as $\sigma_{xy}$. The respective quantities for the second image $y$ ($\mu_y$ and $\sigma_y$) are defined analogously. Then, the luminance, contrast, and structure are given by:
\begin{equation}
    l(x,y)=\frac{2 \mu_{x} \mu_{y}}{\mu_{x}^{2}+\mu_{y}^{2}}
    \label{eq_l}
\end{equation}
\begin{equation}
    c(x,y)=\frac{2 \sigma_{x} \sigma_{y}}{\sigma_{x}^{2}+\sigma_{y}^{2}}
    \label{eq_c}
\end{equation}
\begin{equation}
    s(x,y)=\frac{\sigma_{x y}}{\sigma_{x} \sigma_{y}}
    \label{eq_s}
\end{equation}

MS-SSIM (shortened to MS) considers multiple scales $M$ by relaxing the assumption of a fixed image sampling density. Specifically, we compute the MS-SSIM metric as:
\begin{equation}
\label{ms_ssim}
    \textrm{MS}(x, y)= l_M(x, y) \prod_{j=1}^{M} c_{j}(x, y) s_{j}(x, y)
\end{equation}
where $c_j$ and $s_j$ denote the contrast and structure of images $x$ and $y$ at scale $j$ respectively, and $l_M$ denotes the luminance only at scale $M$.
 
\subsection{Information Bottleneck}
Another way to measure ``task-relevance" in random variables is to consider the total amount of information that a compressed (stochastic) representation $Z$ contains about the input variable $X$ and the output variable $Y$. In particular, (variational) information bottleneck (IB) \citep{tishby2000information,chechik2005information,alemi2016deep} is a framework which utilizes MI to quantify this dependence via the following objective:
\begin{equation}
\label{vib}
    \min_{P(Z|X)} I(X;Z) - \beta I(Z;Y)
\end{equation}
where $\beta>0$ controls the importance of obtaining good performance on the downstream task. Given two random variables $X$ and $Z$, $I(X;Z)$ is computed as
\begin{equation}
    D_\textrm{KL}(P(X,Z)||P(X)P(Z)) = \mathbb{E}_{P(X,Z)}\left[ \log \frac{P(X,Z)}{P(X)P(Z)} \right]
\end{equation}

% \looseness=-1
We conjecture that because MS-SSIM is well-correlated with MI \cite{belghazi2018mine}, inputs transformed by CIM should help MI-based representation learning methods such as VIB learn more task-relevant features; we further explore and empirically verify this hypothesis in Section~\ref{sec:experiments}.

\section{Contrastive Input Morphing}
\label{method}

We propose to approximate the information content between task-relevant and irrelevant features via local, pixel-level image statistics
learned through a triplet loss. Our goal is to leverage both these statistics and labeled examples to \textit{learn} desirable input-space transformations of the data for improved performance on various downstream tasks.

\subsection{Transformation Network Training Procedure}
\paragraph{Network Architecture and Input Transformation.} We utilize a convolutional autoencoder for the Transformation Network (TN) to learn the appropriate transformation for each data point. 
% \kristy{review architecture} 
The TN, which is parameterized by $\phi$, takes in an image $x \in \mathbb{R}^{H\times W \times C}$ and produces 
% intermediate feature activations 
% (weight matrix) 
a weight matrix 
$m \in \mathbb{R}^{H\times W \times 1}$ normalized by the sigmoid activation function, where $H \times W$ denotes the height and width of the image, and $C$ denotes the number of channels. We then use this weight matrix $m$ to transform the input samples by composing it with the learned mask via element-wise multiplication, which gives us the final transformed image $\psi(x) = m \odot x$. The classifier $f_\theta(\cdot)$ is trained via the usual cross entropy loss on $\psi(x)$. 

\paragraph{Sampling Positive and Negative Examples.} We note that there exist several strategies for sampling triplets in Eq.~\ref{eq:ssim}. In CIM, we independently sample one $\xpos$ and one $\xneg$ for each $x$: that is, for each minibatch of anchor points $\{x_i\}_{i=1}^n$ of size $n$, we sample $n$ distinct positive examples and $n$ distinct negative examples during training (for a total minibatch size of $3n$). We further explore the effect of the sampling procedure on CIM's performance in Section~\ref{ablations}.

\paragraph{Triplet Loss.} 
The TN is trained via a triplet loss that operates over sets of three examples at a time: $(x, \xpos, \xneg)$, where given an anchor point $x$, $\xpos$ is a positive example from the same class as $x$, and $\xneg$ is a 
negative example from a different class as $x$. To encourage the transformed input's pixel-wise statistics to be more similar to those of the positive examples (while remaining more dissimilar from those of the negative examples), we apply the MS-SSIM metric from Eq.~\ref{ms_ssim}. Therefore, our triplet loss is defined as:
\begin{equation}
\label{eq:ssim}
\mathcal{L}_{\textrm{con}}(\phi) = \min_\phi \textrm{MS} (\psi(x), x_+) - \\
\textrm{MS}(\psi(x), x_-)
\end{equation} 

\paragraph{Learning Objective.} Therefore, our final objective is:
% The final loss function for the classification problem can be written as:
\begin{equation}
    \label{cim_obj}
    \mathcal{L}_\textrm{CIM}(\phi, \theta) = \lambda\mathcal{L}_{\textrm{con}}(\phi) + \mathcal{L}_{\textrm{sup}}(\theta)
\end{equation}
where 
$\lambda > 0$ is a hyperparameter which controls the contribution of the triplet loss from Eq.~\ref{eq:ssim}, and $\mathcal{L}_{\textrm{sup}}(\theta)$ is the standard cross entropy loss for multi-class classification using the classifier $f_\theta$. The parameters for the transformation network $(\phi)$ and the classifier $(\theta)$ are trained jointly. 

\subsection{Additional Variants of CIM}
\label{addtl-variants}
We further explore different variants of CIM based on the observation that there exist various strategies for measuring task-relevant information. For brevity, we report the results of the additional variants in Appendix~\ref{more_results}.

\paragraph{CIM\textsubscript{g}.} In this setup, we draw inspiration from the neural style transfer literature \cite{gatys2015neural,li2017demystifying,sastry2019detecting} and operate with Gram matrices (inner products) of the triplets' features. Specifically, we modify CIM's loss such that it
% CIM's loss 
encourages 
the positive examples' Gram matrices to move closer together in the embedding space to those of the input, while ensuring that the negative examples' representations are further apart. Therefore the loss is calculated as:
\begin{equation}
\label{eq:cimg}
\mathcal{L}_{\textrm{con}}(\phi) = \min_\phi \{SX_+^T - SX_-^T\}
\end{equation}
where $S$ (Figure~\ref{fig:method}) denotes the learned representation by TN which in our setup has the same dimensions as the input. 

\paragraph{CIM\textsubscript{f}.} To assess whether working in feature space would be more beneficial than working directly in the output space, we also encode the negative and positive samples using the TN in addition to the input.
 
We then create three transformation matrices ($m_1, m_2$, and $m_3$) that we use to modify the input, negative example, and positive example, respectively. These three modified triplets are used to compute the loss during training as in Eq.~\ref{eq:ssim}.

\paragraph{CIM + VIB.}
Finally, we evaluate CIM's compatibility with VIB to demonstrate that our method can be coupled with any (supervised) MI-based representation learning technique. In the CIM+VIB approach, in addition to modifying the input with CIM, we regularize the final feature vector $Z$ of the classifier (the layer before the softmax) with the VIB objective. Therefore, the final loss is given by the following: 
\begin{equation}
    \label{cimvib_obj}
    \mathcal{L}_{\textrm{CIM+VIB}}(\phi, \theta) = \lambda\mathcal{L}_{\textrm{con}}(\phi) + \mathcal{L}_{\textrm{sup}}(\theta) +  D_\textrm{KL}(Q(Z|X)||P(Z))
\end{equation}
where $P(Z) \sim \mathcal{N}(0,I)$ denotes the prior distribution over the stochastic representations $Z$ and $Q(Z|X)$ denotes the approximate posterior over the latent variables. We refer the reader to Appendix~\ref{exp_details} for more details on the VIB architecture and hyperparameters. %\kristy{fix reference}

\section{Experimental Results}
\label{sec:experiments}
For our experiments, we are interested in empirically investigating the following questions:
\begin{enumerate}
    \itemsep0em
    \item Are CIM's learned representations robust to spurious correlations in the input features?
    \item Does the input transformation learned by \modelname~improve domain generalization?
    \item How well can \modelname~preserve classification accuracy across subgroups?

\end{enumerate}

\paragraph{Datasets:} We consider various datasets and tasks to test the effectiveness of our method. We first construct a colored variant of MNIST \citep{lecun1998mnist} to demonstrate that \modelname~successfully ignores nuisance background information in a digit classification task, then further explore this finding on the Background Challenge \citep{xiao2020noise} -- a more challenging dataset.
Next, we evaluate \modelname~on the VLCS dataset \citep{torralba2011unbiased} to demonstrate that the input transformations help in learning representations that generalize to OOD distributions.
Then, we study two benchmark datasets, CelebA \citep{liu2015faceattributes} and Waterbirds \citep{wah2011caltech,zhou2017places}, to show that \modelname~preserves subgroup accuracies.

\begin{figure*}[t!]
  \centering
  \subcaptionbox{\label{ab:subfig1} Visualization of the learned masks.}
    {\includegraphics[width=0.45\textwidth]{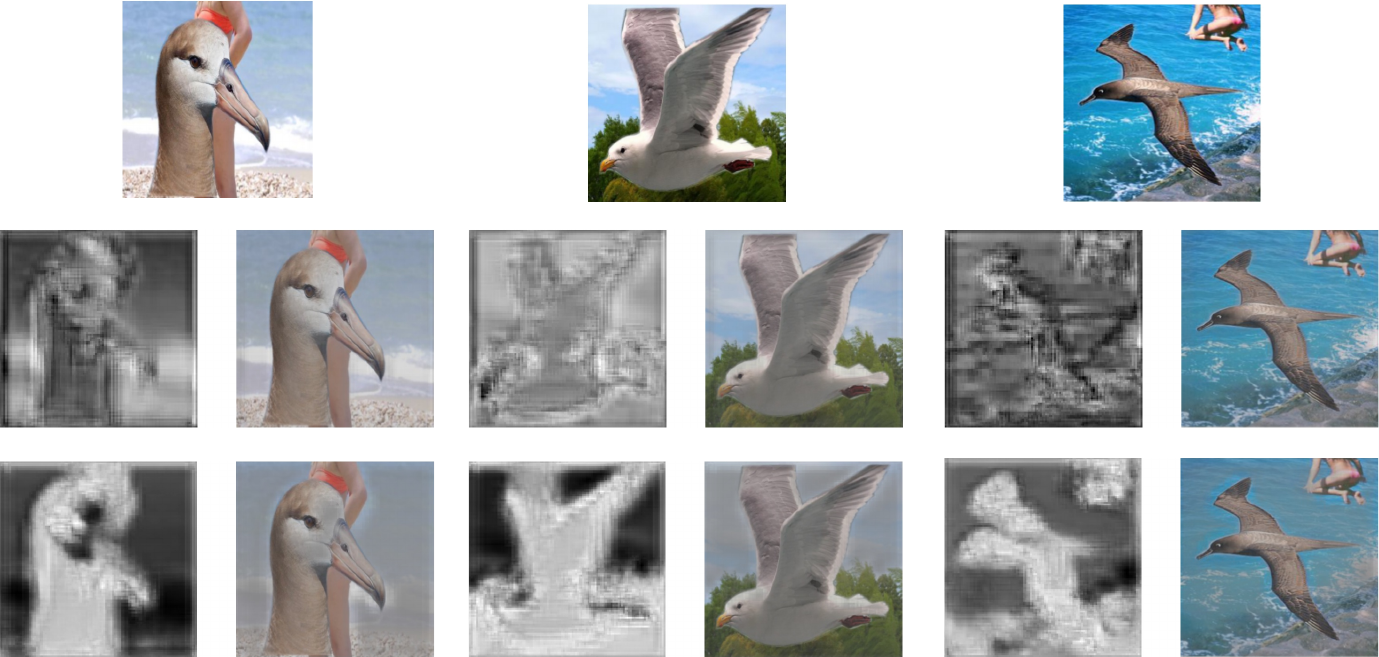}}
  \subcaptionbox{\label{ab:subfig2} Test accuracy in Colored MNIST experiments.}
    {\includegraphics[width=0.5\textwidth]{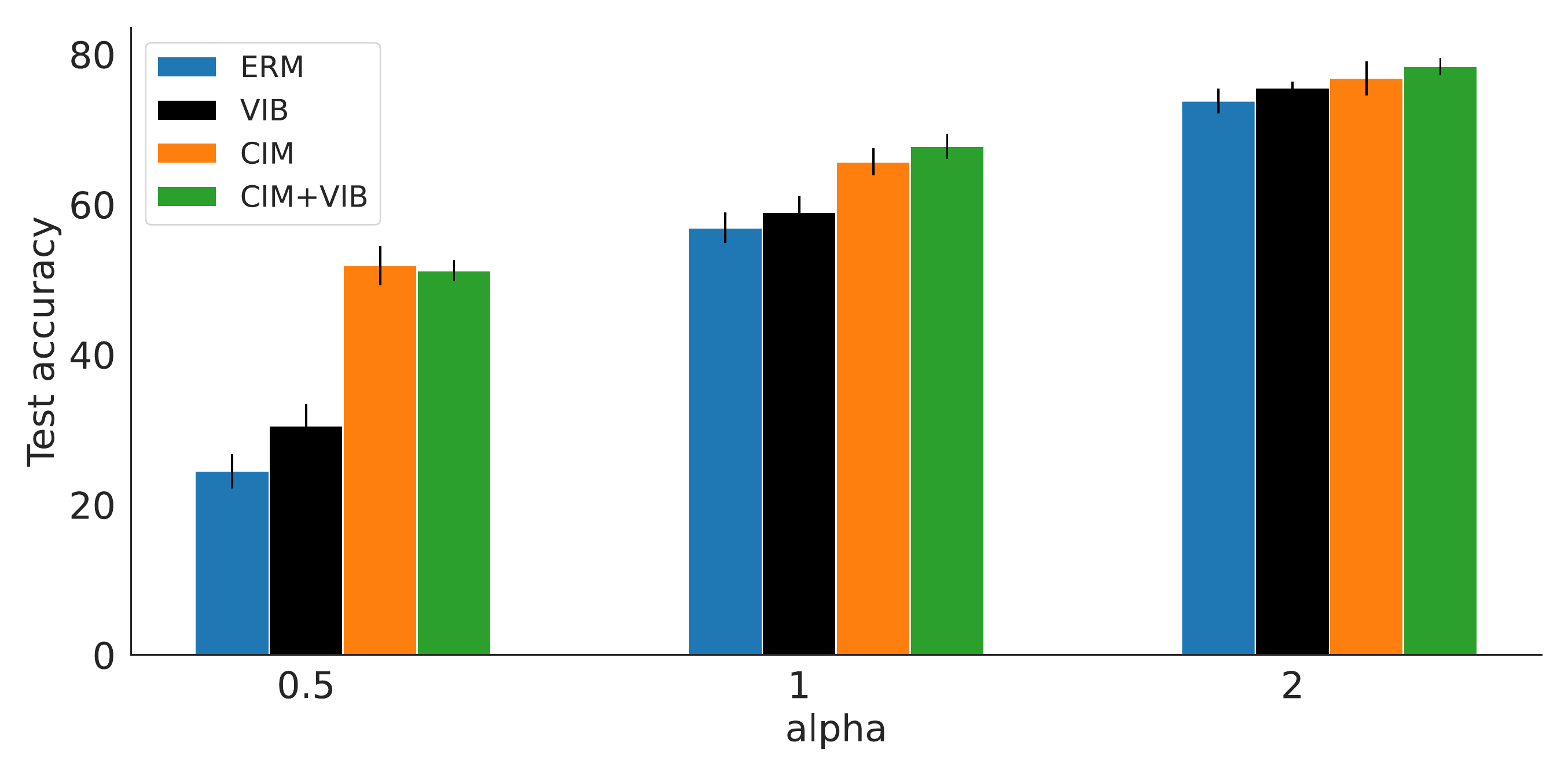}}
  \caption{(a) Qualitative visualizations of the learned masks without and with CIM. The first, second, and third rows show the input, without CIM and with CIM weighting, respectively. The grayscale images are the learned per-pixel weights for each sample. (b) Colored MNIST classification results with different $\alpha$ values, where $\alpha$ denotes the proportion of digits colored with the opposite background color at test time. Both CIM and CIM+VIB significantly outperform ERM. Results are averaged over 3 runs.}
  \label{fig:mnist_and_birds}
\end{figure*}

\paragraph{Models:} We use different classifier architectures depending on the downstream task. While ResNet-50 is the default choice for most datasets, we use ResNet-18 for a fair comparison with existing OOD generalization benchmarks. For the Colored MNIST experiment, we use a simple 3-layered multi-layer perceptron (MLP). The three fully-connected layers are of size 1024, 512, and 256 with ReLU activations.

We also experiment with VIB \citep{alemi2016deep} as both a competing and complementary approach to \modelname. We use ResNet-50 as the encoder in VIB to be consistent with baselines and competing methods. We test three different settings for VIB, where we: (a) apply KL regularization on the last layer of the encoder.

(b) apply KL regularization on the last layer directly, but add 3 fully connected layers before calculating the cross-entropy loss;
and (c) follow variant (a) after adding a fully connected layer after the last layer of the encoder.
We found that (c) worked the best out of all three configurations, and refer the reader to Appendix~\ref{exp_details_arcs} for additional details. 

As a strong contrastive learning baseline, we lightly modify the supervised contrastive learning (SCL) approach to make it comparable with our setup. That is, we train SCL end-to-end (SCL\textsubscript{E2E}) by learning the self-supervised representations jointly with the downstream classification task of interest. Therefore, the contrastive loss of SCL\textsubscript{E2E} becomes:
\begin{equation}
    \mathcal{L}_{\textrm{con}}(\phi) = \min_\phi ||E(x) - E(x_+)||^2 - ||E(x) - E(x_-)||^2
\end{equation}
where $E(.)$ is the last encoded feature vector of size 2048 in ResNet-50 and the model is parameterized by $\phi$.
We refer the reader to Appendix~\ref{exp_details} for additional experimental details and model hyperparameters.

\subsection{Classification with Nuisance Backgrounds}
\label{sec:vib_and_cim}

\paragraph{Colored MNIST:} As a warm-up, we assess whether \modelname~can distinguish between two MNIST digit groups (0-4 and 5-9) in the presence of a spurious input feature (background color). We construct a dataset such that a classifier will achieve low accuracy by relying on background color. For a given value $\alpha$, we color $\alpha$\% of all digits labeled ``0-4" and ``5-9" in the training set with yellow and blue backgrounds, respectively. The remaining $(100-\alpha)$\% of each of the digits' backgrounds are then painted with the opposite color. We vary this proportion by $\alpha=\{0.5, 1.0, 2.0\}$. At test time, we color all the digits labeled ``0-4" in yellow, while coloring the ``5-9" digits' backgrounds in blue. Figure~\ref{fig:mnist_and_birds}b shows that CIM+VIB is better able to utilize relevant information for the downstream task in comparison to ERM by 26.67\%, 10.84\%, and 4.55\% on models trained with $\alpha=\{0.5, 1.0, 2.0\}$ respectively. This suggests that the input transformations learned by CIM are indeed preserving task-relevant information that can be better leveraged by MI-based representation learning methods such as VIB.

\begin{table*}[h!]
\setlength{\tabcolsep}{5pt}
\renewcommand{\arraystretch}{1.1}
\centering
\begin{tabular}{lccccc}
\toprule
\textbf{Method}        & Caltech & LabelMe & Pascal & Sun   & Average  \\ \midrule
DeepC~\citep{li2018deep}         & 87.47   & 62.06   & 64.93  & 61.51 & 68.89 \\
CIDDG~\citep{li2018deep}         & 88.83   & 63.06   & 64.38  & 62.10 & 69.59 \\
CCSA~\citep{motiian2017unified}          & 92.30   & 62.10   & 67.10  & 59.10 & 70.15 \\
SLRC~\citep{ding2017deep}          & 92.76   & 62.34   & 65.25  & 63.54 & 70.15 \\
TF~\citep{li2017deeper}            & 93.63   & 63.49   & 69.99  & 61.32 & 72.11 \\
MMD-AAE~\citep{li2018domain}       & 94.40   & 62.60   & 67.70  & 64.40 & 72.28 \\
D-SAM~\citep{d2018domain}         & 91.75   & 57.95   & 58.59  & 60.84 & 67.03 \\
%JiGen~\citep{carlucci2019domain}         & 96.93   & 60.90   & 70.62  & 64.30 & 73.19 \\
Shape Bias~\citep{asadi2019towards} & 98.11   & 63.61   & 74.33  & 67.11 & 75.79 \\ \midrule
VIB~\citep{alemi2016deep} & 97.44   & 66.41  & 73.29  & 68.49 & 76.41 \\
SCL\textsubscript{E2E} (Ours) & 95.56   & 66.72  & 73.16  & 65.10 & 75.14 \\
\modelname~(Ours)          & 98.21   & \textbf{67.80}  & 73.97  & 69.01 & 77.25 \\
% PM + VIB (Ours) & 97.96   & \textbf{67.06} &  74.35 & 69.26 & 77.16 \\
% CIMg + VIB (Ours) & 98.04   & \textbf{67.10}  & 74.82 & 68.94 & 77.23  \\
CIM + VIB (Ours) & \textbf{98.81}   & 66.49  & \textbf{74.89}  & \textbf{70.13} & \textbf{77.58} \\
\bottomrule
\end{tabular}
\caption{Multi-source domain generalization results (\%) on the VLCS dataset with ResNet-18 as the base network for downstream classification. All reported numbers are averaged over three runs. CIM+VIB outperforms the state-of-the-art model~\citep{asadi2019towards}.}
\label{tab:ood}
\end{table*}

\textbf{Background Challenge:} To evaluate whether the favorable results from Colored MNIST translate to a more challenging setup, we test \modelname~on the Background Challenge~\citep{xiao2020noise}. The Background Challenge is a public dataset consisting of ImageNet-9 \citep{deng2009imagenet} test sets with varying amounts of foreground and background signals, designed to measure the extent to which deep classifiers rely on spurious features for image classification. We compare CIM's performance to relevant baselines in 2 test set configurations: \texttt{Mixed-rand} (\texttt{MR}), where the foreground is overlaid onto a random background; and \texttt{Mixed-same} (\texttt{MS}), where the foreground is placed on a background from the same class. The background gap (\texttt{BGp}), or the difference between these two scenarios,
% The gap between these two scenar ios, named the background gap (\texttt{BGp}), 
is a measure of average robustness to varying backgrounds from different image sources.

As shown in Table~\ref{tab:bg}, CIM+VIB outperforms the baseline ResNet-50's performance by 6.6\% on \texttt{Mixed-rand} (\texttt{MR}), 1.6\% on the original test set, and 0.3\% on \texttt{Mixed-same} (\texttt{MS}). More importantly, CIM+VIB reduces the background gap (\texttt{BGp}) by 6.3\% as compared to the baseline ResNet-50, and 1.4\% as compared to VIB alone.
These results demonstrate that our method indeed learns task-relevant representations without relying on nuisance sources of signal. 
We note that although SCL\textsubscript{E2E} achieves slightly higher accuracy on the original test set, the background gap (\texttt{BGp}) remains relatively large, which is undesirable. Additionally, as shown in Figure~\ref{ab:subfig2}, SCL\textsubscript{E2E} is more computationally expensive as compared to other methods. 

\setlength{\tabcolsep}{2pt}
\renewcommand{\arraystretch}{1}
\begin{table}[ht!]
\centering
\begin{tabular}{lcccc}
\toprule
& \texttt{OR} ($\uparrow$)  & \texttt{MS} ($\uparrow$)  & \texttt{MR} ($\uparrow$)  & \texttt{BGp} ($\downarrow$)   \\ \midrule
% \textcolor{gray}{AlexNet~\citep{xiao2020noise} } & \textcolor{gray}{86.7}   & \textcolor{gray}{76.2}  &  \textcolor{gray}{54.2} & \textcolor{gray}{22.0}  \\
% \textcolor{gray}{ShufleNet~\citep{xiao2020noise} } & \textcolor{gray}{95.7}   & \textcolor{gray}{86.7}  &  \textcolor{gray}{69.4} & \textcolor{gray}{17.3}  \\
% \textcolor{gray}{VGG16~\citep{xiao2020noise} } & \textcolor{gray}{97.6}   & \textcolor{gray}{91.0}  &  \textcolor{gray}{78.0} & \textcolor{gray}{13.0}  \\
% \textcolor{gray}{Res50wx2~\citep{xiao2020noise} } & \textcolor{gray}{97.2}   & \textcolor{gray}{90.6}  &  \textcolor{gray}{78.0} & \textcolor{gray}{12.6}  \\
Res50~\citep{xiao2020noise}   & 96.3   & 89.9  &  75.6 & 14.3 \\ \midrule
VIB~\citep{alemi2016deep} & 97.4   & 89.9  &  80.5 & 9.4 \\
\modelname~(Ours) & 97.7   & 89.8  &  81.1 & 8.8  \\
SCL\textsubscript{E2E} (Ours) & \textbf{98.2}   & \textbf{90.7}  &  80.1 & 10.6 \\
% CIMg + VIB (Ours) & 97.9  & 90.1  &  81.9 & 8.2  \\ 
CIM + VIB (Ours) & 97.9   & 90.2  &  \textbf{82.2} & \textbf{8.0}  \\ 
% PM + VIB (Ours) & \textbf{98.3}   &  \textbf{91.4}  & \textbf{82.3}  &  9.1 \\ 
\bottomrule
\end{tabular}
\caption{Results from the Background Challenge on ImageNet-9 using ResNet-50. Our method outperforms the relevant baselines across all three datasets. The difference between \texttt{MS} and \texttt{MR} is the background gap (\texttt{BGp}).
\texttt{OR} corresponds to accuracy on the original test set.}
\label{tab:bg}
\vspace{-0.4cm}
\end{table}

\subsection{Out-of-Domain Generalization}
\label{sub:ood}

In this experiment, we evaluate \modelname~on OOD generalization performance using the VLCS benchmark dataset ~\citep{torralba2011unbiased}. VLCS consists of images from five object categories shared by the PASCAL VOC 2007, LabelMe, Caltech, and Sun datasets, which are considered to be four separate domains. We follow the standard evaluation strategy used in~\citep{carlucci2019domain}, where we partition each domain into a train (70\%) and test set (30\%) by random selection from the overall dataset. We use ResNet-18 as the backbone to make a fair comparison with the state-of-the-art. As summarized in Table~\ref{tab:ood}, CIM+VIB outperforms the state-of-the-art (Shape Bias~\citep{asadi2019towards}) on each domain and by 1.79\% on average, bolstering our claim that using a lossy transformation of the input is helpful for learning task-relevant representations that generalize across domains.

\subsection{Preservation of Subgroup Performance}\label{sub:subgroup}
We investigate whether representations learned by \modelname~perform well on all subgroups on CelebA and Waterbirds datasets. Preserving good subgroup-level accuracy is challenging for naive ERM-based methods, given their tendency to latch onto spurious correlations \citep{kim2019learning,arjovsky2019invariant,chen2020self}. Most prior works leverage privileged information such as group labels to mitigate this effect \citep{ben2013robust,vapnik2015learning,sagawa2019distributionally,goel2020model,xiao2020noise}. As the TN is trained to capture task-relevant features and minimize nuisance correlations between classes, we hypothesize that \modelname~should perform well at the subgroup level \textit{even without explicit subgroup label information}. 

For a fair comparison, we use ResNet-50 as the backbone in all of our trained models. Table~\ref{tab:celeb-water} shows that CIM+VIB outperforms unsupervised methods on CelebA and Waterbirds in terms of worst-group accuracy, while significantly improving over ERM by 42.49\% and 17.23\% on CelebA and Waterbirds datasets, respectively. We emphasize that the favorable performance of CIM+VIB is obtained \textit{without using subgroup labels}, in contrast with supervised approaches. For additional results with different variants of our method, we refer the reader to Appendix~\ref{more_results}.

\begin{table*}[t!]
\setlength{\tabcolsep}{4pt}
\renewcommand{\arraystretch}{1.1}
\centering
\begin{tabular}{clccc}
\toprule
\textbf{Dataset}                     & \multicolumn{1}{c}{\textbf{Method}} & Unsupervised (subgroub-level) & Worst group acc. & Average acc. \\ \toprule
\multirow{5}{*}{\STAB{\rotatebox[origin=c]{90}{CelebA}}}     
                            & GDRO~\citep{sagawa2019distributionally}                      & \xmark                       & \textbf{88.30}                & 91.80             \\ %\cmidrule{2-5}
                            & ERM                                 & \cmark             &  41.10             & 94.80            \\
                            & Baseline (Ours)                                                 & \cmark                       & 70.31               & 93.98             \\
                            & SCL\textsubscript{E2E} (Ours)                                & \cmark                       & 68.80                & \textbf{95.80}          \\
                            & VIB~\citep{alemi2016deep}                                    & \cmark                       & 78.13               & 91.94          \\
                            % & InfoMask~\citep{taghanaki2019infomask}                       & \cmark                       &                 &           \\
                            % & PM (Ours)                                      & \cmark              &   75.78           &     91.94        \\
                            % & CIMg (Ours)                                      & \cmark              &  82.81            &  89.75           \\ 
                            & CIM (Ours)                                      & \cmark              &  81.25            &   89.24          \\ 
                            % & PM + VIB (Ours)                                      & \cmark              &   82.03           &    92.01         \\ 
                            % & CIMg + VIB (Ours)                                      & \cmark              &  82.03            &  91.27           \\ 
                            & CIM + VIB (Ours)                                      & \cmark              & \underline{83.59}             &  90.61           \\ \midrule

\multirow{6}{*}{\STAB{\rotatebox[origin=c]{90}{Waterbirds}}} 
                            & GDRO~\citep{sagawa2019distributionally}                      & \xmark                       & 83.80                & 89.40             \\
                            & CAMEL~\citep{goel2020model}                                  & \xmark              & \textbf{89.70}                & 90.90             \\ %\cmidrule{2-5}
                            & ERM                                 & \cmark                       & 60.00                & \textbf{97.30}             \\
                            & Baseline (Ours)                                                 & \cmark                       & 62.19               & 96.42             \\
                            & SCL\textsubscript{E2E} (Ours)                                & \cmark                       & 64.10                & 96.50          \\
                            & VIB~\citep{alemi2016deep}                                    & \cmark                       & 75.31               & 95.39          \\
                            % & InfoMask~\citep{taghanaki2019infomask}                       & \cmark                       & 58.4                & 94.9           \\
                            % & PM (Ours)                                      & \cmark              &   68.12          &    95.91         \\
                            % & CIMg (Ours)                                      & \cmark              & 68.75             & 95.26            \\ 
                            & CIM (Ours)                                      & \cmark              & 73.35             & 89.78            \\ 
                            % & PM + VIB (Ours)                                      & \cmark              & 77.19            &  95.71           \\ 
                            % & CIMg + VIB (Ours)                                      & \cmark              & 73.79             &  94.91           \\ 
                            & CIM + VIB (Ours)                                      & \cmark              &  \underline{77.23}            &  95.60           \\ \bottomrule
\end{tabular}
\caption{Average and worst-group accuracies for CelebA and Waterbird benchmark datasets. Methods without group-level supervision (\cmark) are preferable over those with group-level supervision (\xmark). CIMs~+~VIB outperforms unsupervised methods on both datasets, while achieving comparable performance against supervised approaches.  Underline shows the best accuracy among the unsupervised methods.}
\label{tab:celeb-water}
\end{table*}

\subsection{Ablation Studies}
\label{ablations}
Next, we perform a series of ablation studies to assess the contributions of each component in CIM.

\textbf{Effect of the contrastive loss.}
First, we investigate how much of the performance improvement in our method is due to the transformation learned via the \textit{contrastive loss}, rather than a general attention-like mechanism that operates directly over a single input (without any positive or negative examples). As shown in Figure~\ref{ab:subfig1}, we find that simply learning a reweighting matrix $m$ via the TN with $\lambda=0$ in Eq.~\ref{cim_obj} leads to a significant performance degradation relative to CIM on the Waterbirds dataset. For a more in-depth qualitative analysis, we visualize the learned transformations both with and without CIM after they have been composed with the input. As shown in Figure~\ref{fig:mnist_and_birds}a, we observe that the the transformation learned by CIM places less emphasis on the task-irrelevant information (background).

\textbf{Effect of coupling VIB and CIM.} Next, we analyze how much of the performance improvement of CIM+VIB over relevant baselines is from the learned input transformation, rather than using VIB as a downstream feature extractor. As shown in Table~\ref{tab:ablation1}, we observe that CIM significantly improves VIB's performance (9.01\%, 1.92\%, 1.17\% improvements on Colored MNIST, Waterbirds, and VLCS respectively), demonstrating that the learned transformation is indeed useful for better representation learning.

\begin{table}[h!]
\setlength{\tabcolsep}{6pt}
\renewcommand{\arraystretch}{1.15}
\centering
\begin{tabular}{lccc}
\toprule
Method        & Colored MNIST & Waterbirds   & VLCS \\ \midrule
Baseline &   54.87    &  62.19 &    75.83    \\
VIB     & 56.82     & 75.31 &     76.41     \\
% PM      & 67.50      & 76.42 & 69.63        \\
% PM+VIB  & 75.16      & 77.15 & 70.43        \\
CIM     &  65.27     & 73.34  &     77.17     \\
CIM+VIB &  \textbf{65.83}     & \textbf{77.23} &     \textbf{77.58}    \\ \bottomrule
\end{tabular}
\caption{Ablation study of CIM+VIB. Higher is better. We find that the learned input-space transformation helps VIB extract more predictive features for the downstream classification task of interest, as CIM+VIB outperformed all model configurations.}
\label{tab:ablation1}
\end{table}

\textbf{Effect of positive and negative samples.} Additionally, we evaluate the effect of the positive and negative terms in our contrastive loss function on the Waterbirds dataset. Table~\ref{tab:posneg} demonstrates that both the negative and positive samples contribute to improving the worst-group accuracies. In particular, the worst- and second worst-group accuracies improve by 5.07\% and 5.48\% respectively, relative to the model without the contrastive loss (i.e. only $m$). We present additional results on sampling strategies for the positive and negative examples in Table~\ref{tab:add_results} of the Appendix.

\begin{table}[h!]
\setlength{\tabcolsep}{3pt}
\renewcommand{\arraystretch}{1.15}
\centering
\begin{tabular}{lccc}
\toprule
Method                 & Worst-group & 2nd worst-group  \\ \midrule
Only $m$ (without CIM) & 68.28            & 68.75                        \\
CIM (only $x_-$)        & 69.29            & 71.25                        \\
CIM (only $x_+$)        & 72.81            & 73.59                       \\
CIM (both $x_+$ and $x_-$) & \textbf{73.35}            & \textbf{73.75}                       \\ \bottomrule
\end{tabular}
\caption{Ablation study for the effect of positive and negative samples on the worst-group and second worst-group accuracy on Waterbirds dataset. The triplet loss, which leverages both positive and negative samples, outperformed the other configurations.}
\label{tab:posneg}
\end{table}

\begin{figure}[h!]
  \centering
  \subcaptionbox{\label{ab:subfig1} Effect of CIM on accuracy.}
    % {\includegraphics[width=.49\linewidth]{figures/mnist_ab_.pdf}} %\hspace{0.1mm} %\qquad
    {\includegraphics[width=.49\linewidth]{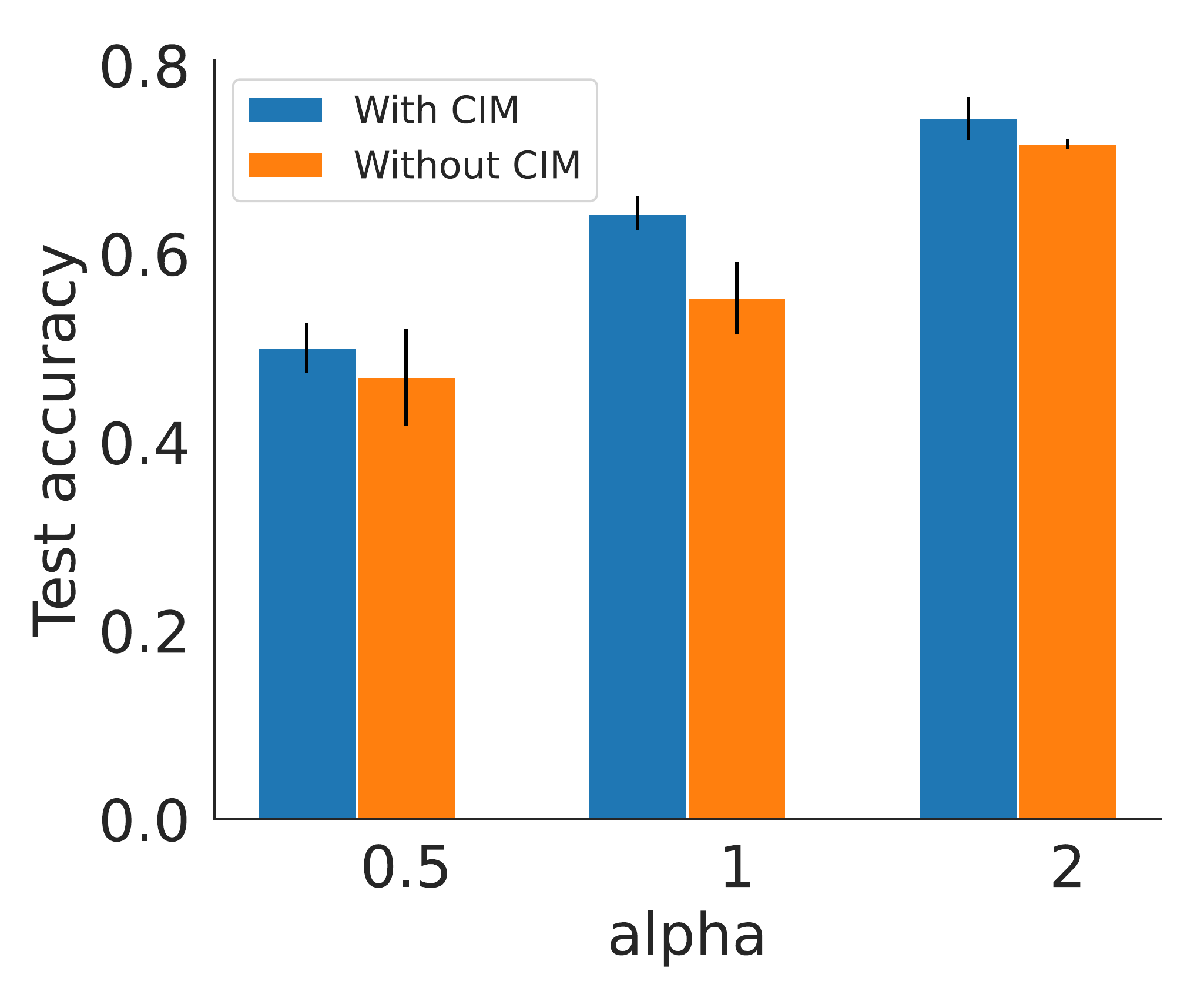}} 
  \subcaptionbox{\label{ab:subfig2} Wall clock time per epoch.}
    % {\includegraphics[width=.5\linewidth]{figures/times_.pdf}}
    {\includegraphics[width=.5\linewidth]{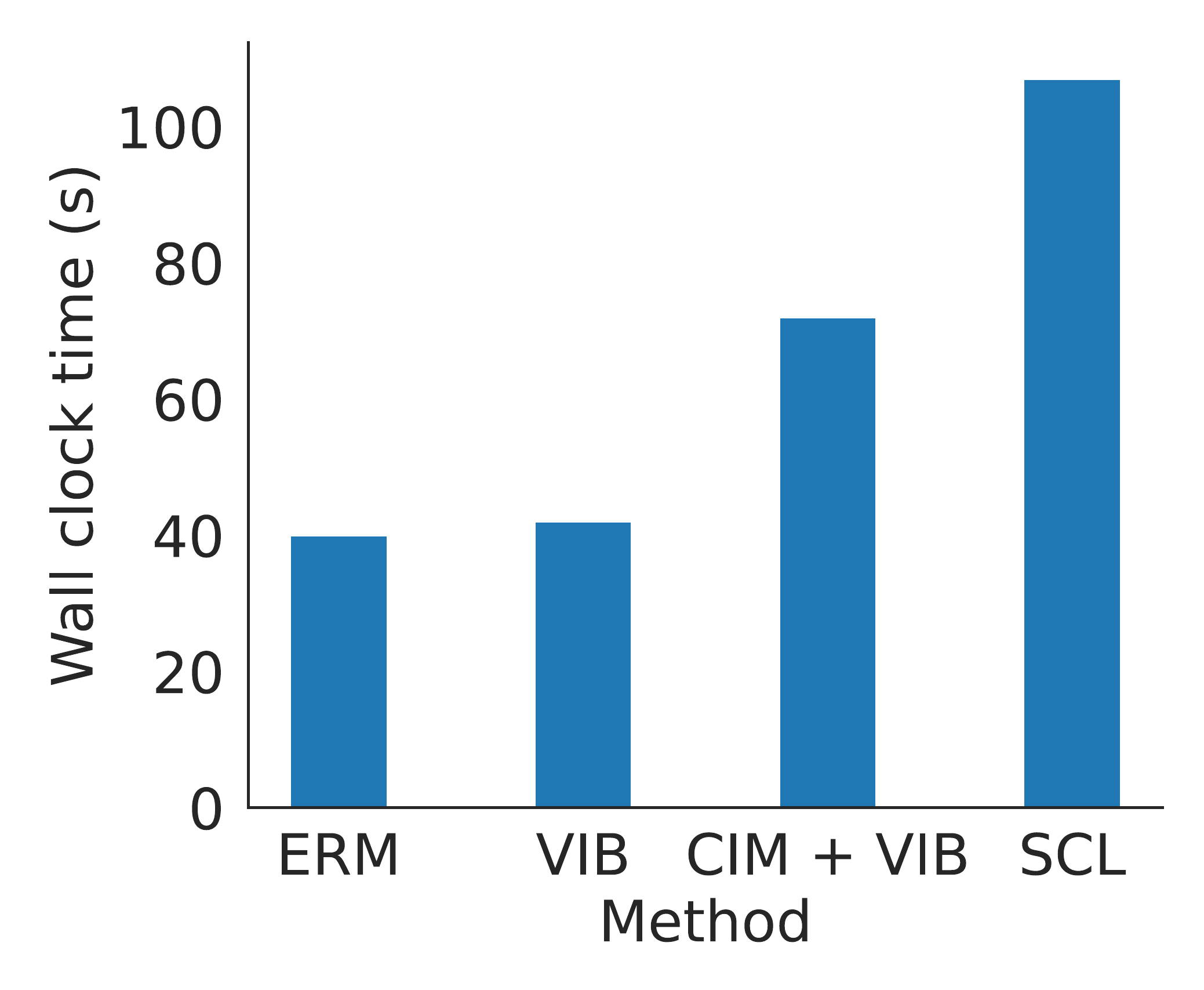}}
  \caption{a) Ablation study investigating the effect of CIM, where we find that simply learning a reweighting matrix without a contrastive loss leads to worse performance. b) Training times per epoch on Waterbirds dataset. SCL is much slower than CIM+VIB ,as it requires encoding all $x, x_-,$ and $x_+$, while CIM+VIB only needs to encode $x$.}
  \label{fig:ab_time}
\end{figure}
\textbf{Computational Overhead.} Finally, we compare the computational overhead of our approach relative to other baselines on the Waterbirds dataset in Figure~\ref{ab:subfig2}. Because CIM needs to calculate the triplet loss for every data point in a minibatch, our method is indeed slower than ERM or VIB. However, we note that we are much faster than SCL, as CIM only requires encoding a single input $x$ via the TN rather than SCL, which must encode all triplets $x$, $x_+$, and $x_-$.

\section{Related Work}
\label{related}
\paragraph{Contrastive representation learning.}
There has been a flurry of recent work in contrastive methods for representation learning, which encourages an encoder network to map ``positive" examples closer together in a latent embedding space while spreading the ``negative" examples further apart \citep{oord2018representation,hjelm2018learning,wu2018unsupervised,tian2019contrastive,arora2019theoretical,chen2020simple}. Some representative approaches include triplet-based losses \citep{schroff2015facenet,koch2015siamese} and variants of noise contrastive estimation \citep{gutmann2010noise}.
In particular, recent work \citep{tian2020makes,wu2020mutual} has shown that \textit{minimizing} MI between views while maximizing predictive information of the representations with respect to the downstream task, leads to performance improvements, similar to IB \citep{chechik2003extracting}. While most contrastive approaches are self-supervised, \citep{khosla2020supervised} utilizes class labels as part of their learning procedure. We emphasize that CIM is not meant to be directly comparable to the aforementioned techniques, as our objective is to learn input transformations of the data that are task-relevant with \textit{labeled supervision}, without relying on a two-stage pretraining approach.

\vspace{-0.3cm}
\paragraph{Robustness of representations.}
Several works have considered the problem of learning relevant features that do not rely on spurious correlations with the predictive task \citep{heinze2017conditional,sagawa2020investigation,chen2020self}.
Though \citep{wang2019learning} is similar in spirit to CIM, they utilize gray-level co-occurrence matrices as the spurious (textural) information of the input images, then regress out this information from the trained classifier's output layer. Our method does not solely rely on textural features and can learn any transformation of the input space that is relevant for the downstream task of interest. Although CIM also bears resemblance to InfoMask \citep{taghanaki2019infomask}, our method is not limited to attention maps. \citep{kim2019learning} uses an MI-based objective to minimize the effect of spurious features, while \citep{pensia2020extracting} additionally incorporates regularization via Fisher information to enforce robustness of the features. In contrast, CIM uses an orthogonal approach to learn robust representations via the perceptual similarity metric in input space.
\vspace{-0.4cm}

\paragraph{SSIM-based loss functions.}
SSIM and MS-SSIM \cite{wang2003multiscale,wang2004image} have seen a recent resurgence in neural network-based approaches. In particular, \cite{yang2020net} adapted SSIM for single image dehazing and signal reconstruction, while \cite{lu2019level} and \cite{snell2017learning} utilized the metric for image generation and downstream classification. Additionally, \citet{abobakr2019ssimlayer} proposed a SSIM layer to make convolutional neural networks robust to noise and adversarial attacks. Though similar in spirit, CIM uses a triplet loss based on SSIM in order to learn input-space transformations of images for robust representations.

\section{Conclusion}
\label{conclusion}
In this work, we considered the problem of extracting representations with  \textit{task-relevant} information from high-dimensional data.
We introduced a new framework, CIM, which learns input-space transformations of the data via a triplet loss to mitigate the effect of irrelevant input features on downstream classification performance. Through experiments on (1) classification with nuisance background information; (2) out-of-domain generalization; and (3) preservation of subgroup performance, which typically suffer from the presence of spurious correlations in the data, we showed that CIM outperforms most relevant baselines. Additionally, we demonstrated that CIM is complementary to other mutual information-based representation learning frameworks such as VIB. 
A limitation of our work is precisely the need for labeled supervision, which may be difficult or prohibitively expensive to obtain during training.
For future work, it would be interesting to test different types of distance metrics for the triplet loss, to explore whether CIM can be used as an effective way to \textit{learn} views for unsupervised contrastive learning, and to investigate label-free approaches for learning the input transformations.

\subsubsection*{\normalsize{Acknowledgements}}
We are thankful to Pang Wei Koh, Shiori Sagawa, and Aditya Sanghi for helpful discussions. KC is supported by the NSF GRFP, Stanford Graduate Fellowship, and Two Sigma Diversity PhD Fellowship.

\bibliography{references}
\bibliographystyle{icml2021}

\raggedbottom
\pagebreak

\onecolumn
\section*{Appendix}
\renewcommand{\thesubsection}{\Alph{subsection}}
\label{appendix}

\subsection{Additional Experimental Details}
\label{exp_details}

\subsubsection{Architectures}
\label{exp_details_arcs}
\begin{figure*}[ht!]
  \centering
  \subcaptionbox{Colored MNIST \label{tn:subfig1}}
    {\includegraphics[width=.2\linewidth]{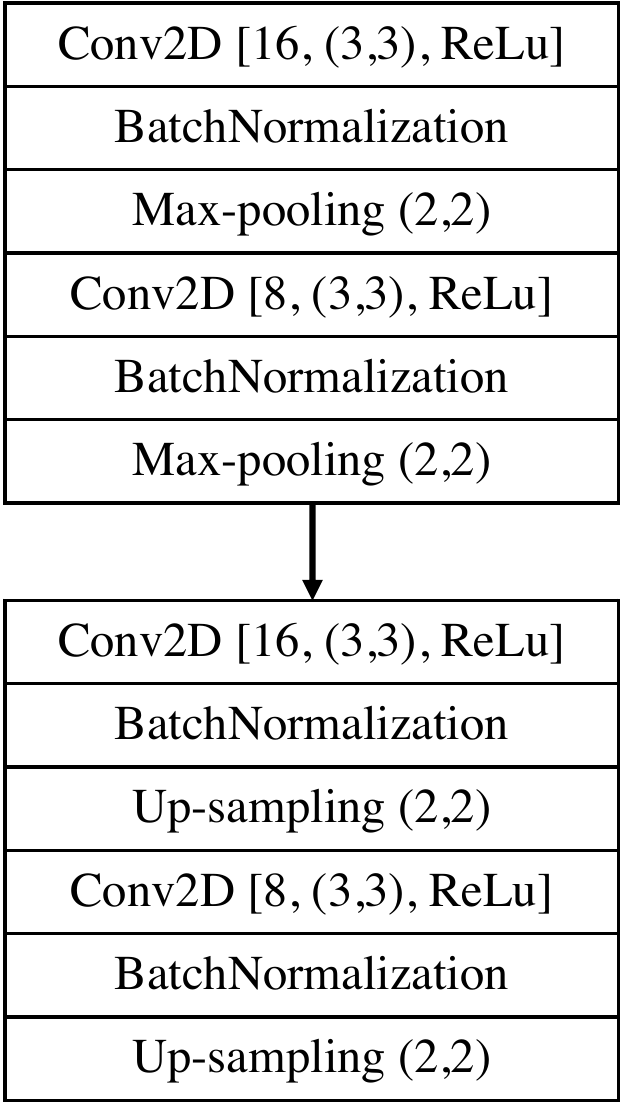}} 
    % \hspace{10.0mm} %\qquad
  \subcaptionbox{Other datasets \label{tn:subfig2}}
    {\includegraphics[width=.22\linewidth]{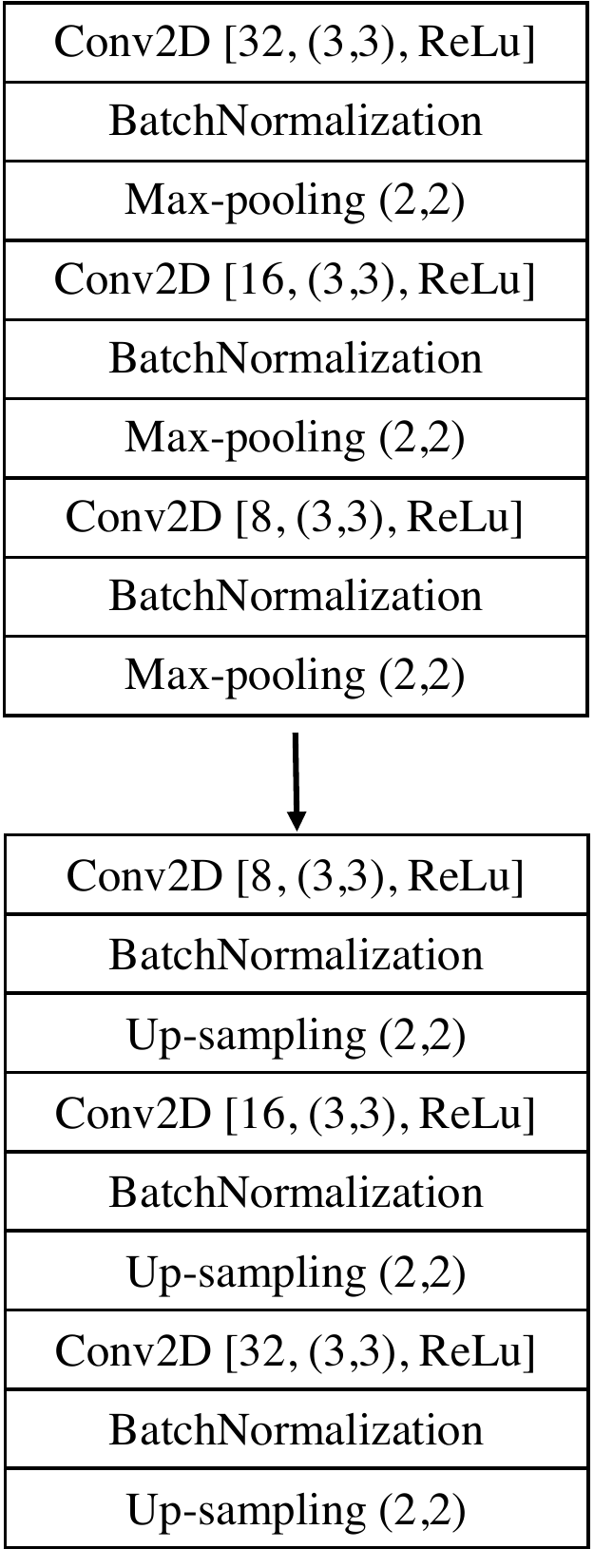}}
  \caption{a) Architecture of the TN used for all experiments in the paper.}
  \label{fig:tnblock}
\end{figure*}

\paragraph{TN Architectures.} In Figure~\ref{fig:tnblock}, we show the detailed TN architectures used for the Colored MNIST experiment, as well as for the other models. Architecture details for the downstream classifiers that are specific to each task can be found in Section~\ref{sec:experiments}.

\paragraph{VIB Architectures.} We tested 3 different approaches where we: (a) apply KL regularization on the last layer of the encoder which is of size (1, 2048) and then compute the cross entropy loss;
(b) apply KL regularization on the last layer directly, but add 3 fully connected layers 
of \texttt{(1024, ReLU, batch normalization)}, \texttt{(512, ReLU, batch normalization)}, and \texttt{(256, ReLU)}, before calculating the cross-entropy loss;
and (c) follow variant (a) after adding a fully connected layer of size 512 after the last layer of the encoder.
We found that (c) yielded the best performance.

\pagebreak
\subsubsection{Hyperparameter Configurations}
\label{hyperparams}
For all the methods including the baselines we have tuned the hyper-parameters such as the optimizer (SGD and Adam), learning rate (0.01, 0.001. 0.0001), and batch size (16, 32, 64). The input size for all the experiments was set to $224 \times 224 \times 3$ except for Colored MNIST, which was $28 \times 28 \times 3$. For Colored MNIST and CelebA, we found that the Adam optimizer with a learning rate of $0.0001$ and batch size of 64 worked best. For the Waterbirds and VLCS datasets, we found that the SGD optimizer with learning rate of $0.001$ and momentum of $0.9$ with batch size of 32 yielded the best performance. For the background challenge, we set the optimizer to SGD with learning rate of $0.001$ and batch size to 32.

For CIM and VIB-based approaches, we tested $\beta$ (the hyperparameter for VIB) and $\lambda$ (for the contrastive loss) with a range of values within $[1.0, 0.00001]$. In Table~\ref{tab:params}, we report the best-performing hyperparameters for each method.

\begin{table}[h!]
\centering
\resizebox{0.5\linewidth}{!}{
\setlength{\tabcolsep}{4pt}
\begin{tabular}{lllc}
\midrule
                                 Dataset      &  Method &  VIB's $\beta$ & CIM's $\lambda$  \\ \midrule
\multirow{3}{*}{CelebA}                & CIM                  & - &  0.00001                    \\ 
                                      & VIB                 & 0.1 &  -                      \\ 
                                      & CIM+VIB           & 0.001 & 0.0001                     \\ \midrule
                                       
\multirow{3}{*}{Waterbirds}           & CIM     & -  &  0.0001                      \\ 
                                      & VIB       & 0.001 & -                      \\ 
                                      & CIM+VIB    & 0.001 & 0.0001                \\ \midrule
                                       
\multirow{1}{*}{Background challenge}  & CIM+VIB     & 0.001 & 0.0001                \\ \midrule

\multirow{3}{*}{Color MNIST}          & CIM                 & - & 0.00001                       \\ 
                                      & VIB                  & 0.00001 & -                        \\ 
                                      & CIM+VIB              & 0.00001 & 0.00001                  \\ \midrule
\end{tabular}
}
\caption{$\beta$ and $\lambda$ values for CIM, VIB, and CIM+VIB.}
\label{tab:params}
\end{table}

% \pagebreak
\subsection{Additional Experimental Results}
\label{more_results}

\subsubsection{Out-of-domain Generalization}
We present additional results from Section~\ref{sub:ood} with standard errors calculated over 3 runs for our methods. We note that the results from other papers did not include replicates. 
\begin{table*}[h!]
\centering
\resizebox{0.8\linewidth}{!}{
\setlength{\tabcolsep}{5pt}
\renewcommand{\arraystretch}{1.1}
\centering
\begin{tabular}{lccccc}
\toprule
\textbf{Method}        & Caltech & LabelMe & Pascal & Sun   & Average  \\ \midrule
DeepC~\citep{li2018deep}         & 87.47   & 62.06   & 64.93  & 61.51 & 68.89 \\
CIDDG~\citep{li2018deep}         & 88.83   & 63.06   & 64.38  & 62.10 & 69.59 \\
CCSA~\citep{motiian2017unified}          & 92.30   & 62.10   & 67.10  & 59.10 & 70.15 \\
SLRC~\citep{ding2017deep}          & 92.76   & 62.34   & 65.25  & 63.54 & 70.15 \\
TF~\citep{li2017deeper}            & 93.63   & 63.49   & 69.99  & 61.32 & 72.11 \\
MMD-AAE~\citep{li2018domain}       & 94.40   & 62.60   & 67.70  & 64.40 & 72.28 \\
D-SAM~\citep{d2018domain}         & 91.75   & 57.95   & 58.59  & 60.84 & 67.03 \\
%JiGen~\citep{carlucci2019domain}         & 96.93   & 60.90   & 70.62  & 64.30 & 73.19 \\
Shape Bias~\citep{asadi2019towards} & 98.11   & 63.61   & 74.33  & 67.11 & 75.79 \\ \midrule
VIB~\citep{alemi2016deep} & 97.44$\pm$0.143  & 66.41$\pm$0.045  & 73.29$\pm$0.040  & 68.49$\pm$0.150 & 76.41$\pm$0.095 \\
SCL\textsubscript{E2E} (Ours) & 95.56$\pm$0.141   & 66.72$\pm$0.043  & 73.16$\pm$0.053  & 65.10$\pm$0.071 & 75.14$\pm$0.077 \\
\modelname~(Ours)          & 98.21$\pm$0.004   & \textbf{67.80$\pm$0.010}  & 73.97$\pm$0.003  & 69.01$\pm$0.003 & 77.25$\pm$0.005 \\
% PM + VIB (Ours) & 97.96   & \textbf{67.06} &  74.35 & 69.26 & 77.16 \\
% CIMg + VIB (Ours) & 98.04   & \textbf{67.10}  & 74.82 & 68.94 & 77.23  \\
CIM + VIB (Ours) & \textbf{98.81$\pm$0.003}   & 66.49$\pm$0.004  & \textbf{74.89$\pm$0.007}  & \textbf{70.13$\pm$0.008} & \textbf{77.58$\pm$0.006} \\
\bottomrule
\end{tabular}
}
\caption{Multi-source domain generalization results (\%) on the VLCS dataset with ResNet-18 as the base network for downstream classification. All reported numbers are averaged over three runs. CIM+VIB outperforms the state-of-the-art model~\citep{asadi2019towards}.}
\label{tab:ood}
\end{table*}

\subsubsection{Additional Variants of CIM on Subgroup Performance}
We present additional results on the other variants of CIM as mentioned in Section 3.2: CIM\textsubscript{f} and CIM\textsubscript{g}. We also experiment with a naive metric based on the $\ell_2$ distance between two images in pixel space, which we name $\textrm{CIM}_{\textrm{mse}}$, and find that it leads to worse performance.
We report the results of both these variants on Waterbirds and CelebA datsets in Table~\ref{tab:celeb-water-app}. We find that encoding only the input (i.e. CIM+VIB) to calculate the structural triplet loss outperforms both CIM\textsubscript{g} and CIM\textsubscript{f}.

\begin{table*}[ht!]
\setlength{\tabcolsep}{2pt}
\centering
\resizebox{0.75\linewidth}{!}{
\begin{tabular}{clccc}
\toprule
\textbf{Dataset}                     & \multicolumn{1}{c}{\textbf{Method}} &  unsupervised (group-level), & Worst-group acc. & Average acc. \\ \toprule
\multirow{8}{*}{\STAB{\rotatebox[origin=c]{90}{CelebA}}}     
                            & GDRO~\citep{sagawa2019distributionally}                      & \xmark                       & \textbf{88.30}                & 91.80             \\ %\cmidrule{2-5}
                            & ERM~\citep{sagawa2019distributionally}                                 & \cmark             &  41.10             & 94.80            \\
                            & Our baseline                                                 & \cmark                       & 70.31               & 93.98             \\
                            & VIB~\citep{alemi2016deep}                                    & \cmark                       & 78.13               & 91.94          \\
                            % & InfoMask~\citep{taghanaki2019infomask}                       & \cmark                       &                 &           \\
                            & SCL\textsubscript{E2E} (Ours)                                & \cmark                       & 68.80                & \textbf{95.80}          \\
                            % & PM (Ours)                                      & \cmark              &   75.78           &     91.94        \\
                            % & CIM\textsubscript{g} (Ours)                                      & \cmark              &  82.81            &  89.75           \\ 
                            & CIM\textsubscript{f} + VIB (Ours)                                      & \cmark              & 80.87            &  88.24           \\ 
                            % & CIMs (Ours)                                      & \cmark              &  81.25            &   89.24          \\ 
                            % & PM + VIB (Ours)                                      & \cmark              &   82.03           &    92.01         \\ 
                            & CIM\textsubscript{g} + VIB (Ours)                                      & \cmark              &  82.03            &  91.27           \\ 
                            & CIM + VIB (Ours)                                      & \cmark              & \underline{83.59}             &  90.61           \\ \midrule

\multirow{9}{*}{\STAB{\rotatebox[origin=c]{90}{Waterbirds}}} 
                            & GDRO~\citep{sagawa2019distributionally}                      & \xmark                       & 83.80                & 89.40             \\
                            & CAMEL~\citep{goel2020model}                                  & \xmark              & \textbf{89.70}                & 90.90             \\ %\cmidrule{2-5}
                            & ERM~\citep{sagawa2019distributionally}                                 & \cmark                       & 60.00                & \textbf{97.30 }            \\
                            & Our baseline                                                 & \cmark                       & 62.19               & 96.42             \\
                            & VIB~\citep{alemi2016deep}                                    & \cmark                       & 75.31               & 95.39          \\
                            % & InfoMask~\citep{taghanaki2019infomask}                       & \cmark                       & 58.4                & 94.9           \\
                            & SCL\textsubscript{E2E} (Ours)                                & \cmark                       & 64.10                & 96.50          \\
                            % & PM (Ours)                                      & \cmark              &   68.12          &    95.91         \\
                            % & CIM\textsubscript{g} (Ours)                                      & \cmark              & 68.75             & 95.26            \\ 
                            & CIM\textsubscript{f} + VIB (Ours)                                      & \cmark              &  76.65            &  95.31           \\ 
                            % & CIMs (Ours)                                      & \cmark              & 73.35             & 89.78            \\ 
                            % & PM + VIB (Ours)                                      & \cmark              & 77.19            &  95.71           \\ 
                            & CIM\textsubscript{g} + VIB (Ours)                                      & \cmark              & 73.79             &  94.91           \\ 
                            & CIM + VIB (Ours)                                      & \cmark              &  \underline{77.23}            &  95.60           \\ \bottomrule
\end{tabular}}
\caption{Average and worst-group accuracies for CelebA and Waterbird benchmark datasets. Methods without group-level supervision (\cmark) are preferable over those with group-level supervision (\xmark). CIMs~+~VIB outperforms unsupervised methods on both datasets, while achieving comparable performance against supervised approaches.  Underline shows the best accuracy among the unsupervised methods.}
\label{tab:celeb-water-app}
\end{table*}

\subsubsection{Connection to Weighted attention and Saliency Maps}
Although our method bears resemblance to methods with (weighted) attention and saliency, we demonstrate that the TN in CIM is in fact learning a more sophisticated transformation. We experimented with two approaches: (1) learning an attention-like map (\texttt{Only m}) via the TN; and (2) GradCAM (Selvaraju et. al 2016) for saliency detection (\texttt{Saliency}) to see which method would yield perform more favorably on the downstream classification task.

\begin{figure*}[h!]
     \centering
     \includegraphics[width=.5\textwidth]{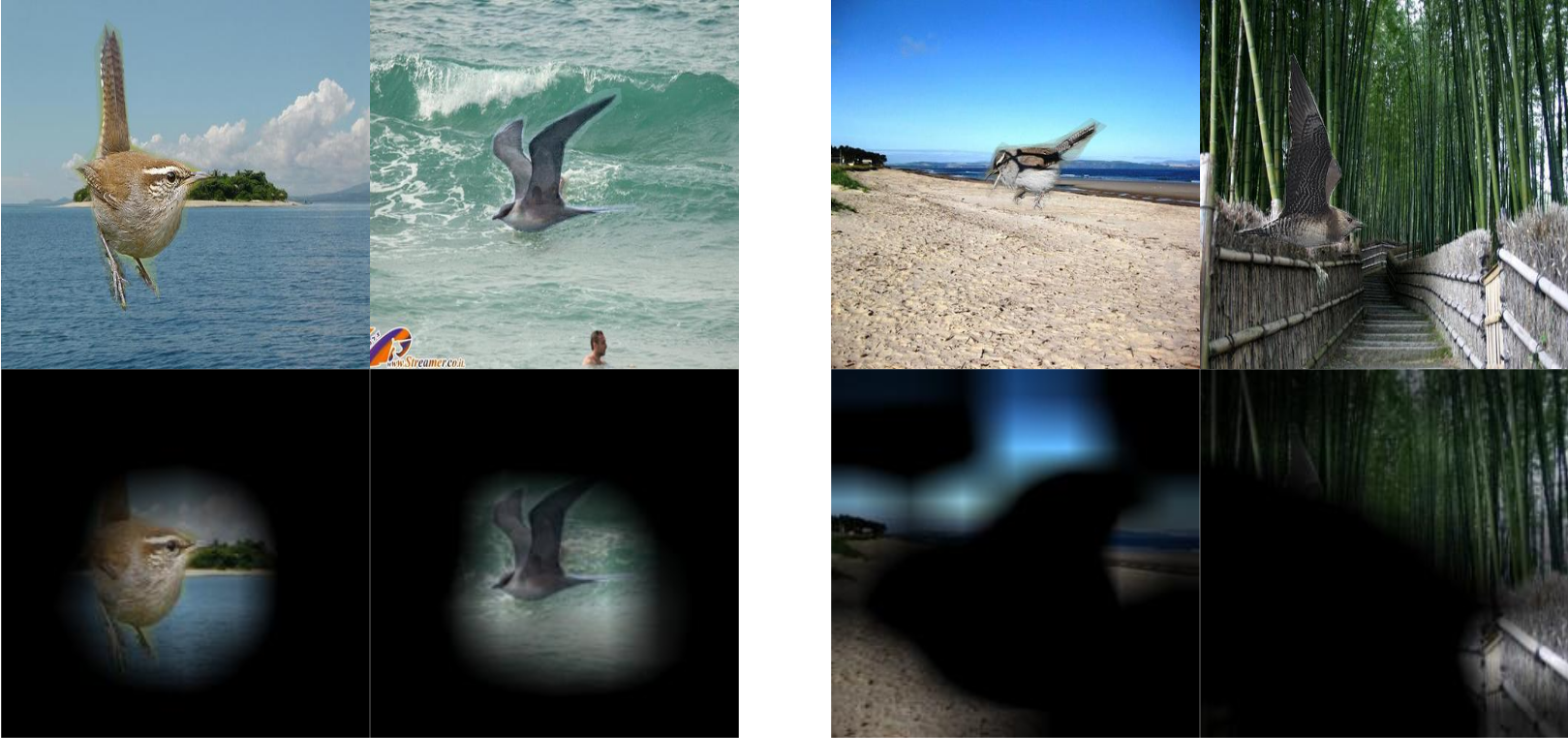}
    %  \vspace{-0.4cm}
     \caption{Learned saliency maps from the GradCAM baseline.}
     \label{fig:saliency}
 \end{figure*}

For the GradCAM baseline, we first trained a ResNet-50 classifier, computed saliency maps using GradCAM after freezing the model, composed the saliency map with the inputs, and trained a new ResNet-50 classifier with the modified inputs (without nuisance background information). As in Table~\ref{tab:add_results}, we find that this saliency method (\texttt{Saliency}) improves over the baseline, but does not outperform either the learned attention map (\texttt{Only m}) or CIM. Figure~\ref{fig:saliency} shows that this is because the saliency detection algorithm may incorrectly mask out a task-relevant region in the original image. By learning this transformation \textit{jointly} with our classifier as in CIM, we are able to improve performance.

\pagebreak
\subsubsection{Alternative sampling strategies for CIM}
Inspired by the contrastive learning literature which explores the impact of various positive and negative sampling strategies, we experimented with two alternative sampling approaches: \texttt{neg++} and \texttt{pos++} (sampling more negative/positive examples per batch respectively). In \texttt{neg++} we used 3 negative examples for each positive example, and vice versa. As shown in Table~\ref{tab:add_results}, we did not find a clear improvement.

\setlength{\tabcolsep}{6pt}
\begin{table*}[h!]
\centering
\resizebox{0.5\linewidth}{!}{
\begin{tabular}{llcccc}
\cline{3-6}
                            &              & \multicolumn{2}{c}{\textbf{CelebA}} & \multicolumn{2}{c}{\textbf{Waterbirds}} \\ \midrule
                    & \textbf{Method}       & \textbf{Worst-group}   & \textbf{Average}   & \textbf{Worst-group}      & \textbf{Average}    \\ \midrule
                            & Baseline & 70.31          & \textbf{93.91}     & 62.19            & \textbf{96.42}       \\ \midrule
\multirow{2}{*}{}         & Only $m$       & 82.81          & 90.06     & 75.31            & 95.14       \\
                            & Saliency     & 75.77          & 90.90     & 67.19            & 92.70       \\ 
                            & CIM\textsubscript{mse}       & 75.78          & 91.55     & 69.69            & 95.46       \\\midrule
\multirow{2}{*}{}         & CIM neg++    & 78.12          & 91.90     & 70.62            & 96.10       \\
                            & CIM pos++    & 78.13          & 92.89     & 76.71            & 95.60       \\ \midrule
                            & CIM          & \textbf{83.59}          & 90.61     & \textbf{77.23}            & 95.60       \\ \bottomrule
\end{tabular}}
% \caption{note that all the methods after second row are +VIB}
\caption{Results for different baselines and sampling strategies.}
\label{tab:add_results}
\end{table*}

\end{document}